\documentclass[10pt,twocolumn,letterpaper]{article}

\usepackage{wacv}
\usepackage{times}
\usepackage{epsfig}
\usepackage{graphicx}
\usepackage{amsmath}
\usepackage{amssymb}
\usepackage{authblk}
\usepackage{multirow}
\usepackage[linesnumbered,ruled]{algorithm2e}
\SetAlFnt{\small}

\def\wacvPaperID{67} 

\wacvfinalcopy 

\ifwacvfinal
\fi

\ifwacvfinal
\usepackage[breaklinks=true,bookmarks=false]{hyperref}
\else
\usepackage[pagebackref=true,breaklinks=true,colorlinks,bookmarks=false]{hyperref}
\fi

\ifwacvfinal
\else
\fi

\begin{document}

\title{Enhancing Diversity in Teacher-Student Networks via Asymmetric branches for Unsupervised Person Re-identification\vspace{-0.8cm}}

\author[1,2]{Hao Chen}
\author[2]{Benoit Lagadec}
\author[1]{Francois Bremond\vspace{-0.5cm}}
\affil[1]{\small University of C\^ote d'Azur, Inria, Stars Project-Team, France\authorcr
  \{\tt\small hao.chen,francois.bremond\}@inria.fr}
\affil[2]{\small European Systems Integration, France\authorcr
  \tt\small benoit.lagadec@esifrance.net\vspace{-0.8cm}}

\maketitle

\begin{abstract}
The objective of unsupervised person re-identification (Re-ID) is to learn discriminative features without labor-intensive identity annotations. State-of-the-art unsupervised Re-ID methods assign pseudo labels to unlabeled images in the target domain and learn from these noisy pseudo labels. Recently introduced Mean Teacher Model is a promising way to mitigate the label noise. However, during the training, self-ensembled teacher-student networks quickly converge to a consensus which leads to a local minimum. We explore the possibility of using an asymmetric structure inside neural network to address this problem. First, asymmetric branches are proposed to extract features in different manners, which enhances the feature diversity in appearance signatures. Then, our proposed cross-branch supervision allows one branch to get supervision from the other branch, which transfers distinct knowledge and enhances the weight diversity between teacher and student networks. Extensive experiments show that our proposed method can significantly surpass the performance of previous work on both unsupervised domain adaptation and fully unsupervised Re-ID tasks. 
\footnote{Code at \url{https://github.com/chenhao2345/ABMT}.}
\end{abstract}

\section{Introduction}
Person re-identification (Re-ID) targets at retrieving a person of interest across non-overlapping cameras. Since there are domain gaps resulting from illumination condition, camera property and view-point variation, a Re-ID model trained on a source domain usually shows a huge performance drop on other domains.

Unsupervised domain adaptation (UDA) targets at shifting the model trained from a source domain with identity annotation to a target domain via learning from unlabeled target images. In the real world, unlabeled images in a target domain can be easily recorded, which is almost labor-free. It is intuitive to use these images to adapt a pretrained Re-ID model to the desired domain. Fully unsupervised Re-ID further minimises the supervision by removing pre-training on the labelled source domain. 

State-of-the-art UDA Person Re-ID methods \cite{ge2020mutual,yang2020asy} and unsupervised methods \cite{Lin2020UnsupervisedPR} assign pseudo labels to unlabeled target images. The generated pseudo labels are generally very noisy. The noise is mainly from several inevitable factors, such as the strong domain gaps and the imperfection of clustering. In this way, an unsupervised Re-ID problem is naturally transferred into Generating pseudo labels and Learning from noisy labels problems.

To generate pseudo labels, the most intuitive way is to use a clustering algorithm, which gives a good starting point for clustering based UDA Re-ID \cite{Zhang2019SelfTrainingWP,Fu2018SelfSimilarityGA}. Recently, Ge \etal \cite{ge2020mutual} propose to add a Mean Teacher \cite{Tarvainen2017MeanTA} model as online soft pseudo label generator, which effectively reduces the error amplification during the training with noisy labels. In this paper, we also use both clustering-based hard labels and teacher-based soft labels in our baseline. 

To handle noisy labels, one of the most popular approaches is to train paired networks so that each network helps to correct its peer, e.g., two-student networks in Co-teaching \cite{Han2018CoteachingRT} and two-teacher-two-student networks in MMT \cite{ge2020mutual}. However, these paired models with identical structure are prone to converge to each other and get stuck in a local minimum. There are several attempts to alleviate this problem, such as Co-teaching+ \cite{Yu2019HowDD}, ACT \cite{yang2020asy} and MMT \cite{ge2020mutual}. These attempts of keeping divergence between paired models are mainly based on either different training sample selection \cite{Yu2019HowDD,yang2020asy} or different initialization and data augmentation\cite{ge2020mutual}. In this paper, we propose a strong alternative by designing asymmetric neural network structure in the Mean Teacher Model. We use two independent branches with different depth and global pooling methods as last layers of a neural network.
Features extracted from both branches are concatenated as the appearance signature, which enhances the feature diversity in the appearance signature and allows to get better clustering-based hard labels. Each branch gets supervision from its peer branch of different structure, which enhances the divergence between paired teacher-student networks. Our proposed decoupling method does not rely on different source domain initializations, which makes it more effective in the fully unsupervised scenario where the source domain is not available.

In summary, our contributions are:
\begin{enumerate}
  \item We propose to enhance the feature diversity inside person Re-ID appearance signatures by splitting last layers of a backbone network into two asymmetric branches, which increases the quality of clustering-based hard labels.
  \item We propose a novel decoupling method where asymmetric branches get cross-branch supervision, which avoids weights in paired teacher-student networks converging to each other and increases the quality of teacher-based soft labels.
  \item Extensive experiments and ablation study are conducted to validate the effectiveness of each proposed component and the whole framework. 
\end{enumerate}

\section{Related Work}
\paragraph{Unsupervised domain adaptive Re-ID.} Recent unsupervised cross-domain Re-ID methods can be roughly categorized into distribution alignment and pseudo label based adaptation. The objective of distribution alignment is to learn domain invariant features. Several attempts \cite{Wang2018TransferableJA,Lin2018MultitaskMF} leverage semantic attributes to align the feature distribution in the latent space. However, these approaches strongly rely on extra attribute annotation, which requires extra labor. Another possibility is to align the feature distribution by transferring labeled source domain images into the style of target domain with generative adversarial networks \cite{wei2018person,Zhong_2018_ECCV,chen2019instance}. 
Style transferred images are usually combined with pseudo label based adaptation to get a better performance. Pseudo label based adaptation is a more straightforward approach for unsupervised cross-domain Re-ID, which directly assigns pseudo labels to unlabelled target images and allows to fine-tune a pre-trained model in a supervised manner. Clustering algorithms are widely used in previous unsupervised cross-domain Re-ID methods. UDAP \cite{song2020unsupervised} provides a good analysis on clustering based adaptation and use a k-reciprocal encoding \cite{zhong2017re} to improve the quality of clusters. PCB-PAST \cite{Zhang2019SelfTrainingWP} simultaneously learns from a ranking-based and clustering-based triplet losses. SSG \cite{Fu2018SelfSimilarityGA} assigns clustering-based pseudo labels to both global and local features. To mitigate the clustering-based label noise, researchers borrow ideas from how unlabeled data is used in Semi-supervised learning and Learning from noisy labels. ECN \cite{zhong2019invariance} uses an exemplar memory to save averaged features to assign soft labels. ACT \cite{yang2020asy} splits the training data into inliers/outliers to enhance the divergence of paired networks in Co-teaching  \cite{Han2018CoteachingRT}. MMT \cite{ge2020mutual} adopts two student and two Mean Teacher networks. Two students are initialized differently from source pre-training in order to enhance the divergence of paired teacher-student networks. Each mean teacher network provides soft labels to supervise peer student network. However, despite different initializations and different data augmentations used in peer networks, the decoupling is not encouraged enough during the training. We directly use asymmetric neural network structure inside teacher-student networks, which encourages the decoupling at all epochs.  

\paragraph{Teacher-Student Network for Semi-Supervised Learning.}
Unsupervised domain adaptation can be regarded to some extent as Semi-Supervised Learning (SSL), since both of them utilize labeled data (source domain for UDA) and large amount of unlabeled data (target domain for UDA). A teacher-student structure is commonly used in SSL. This structure allows student network to gradually exploit data with perturbations under consistency constraints. In $\Pi$ model and Temporal ensembling \cite{Laine2016TemporalEF}, the student learns from either samples forwarded twice with different noise or exponential moving averaged (EMA) predictions under consistency constraints. Instead of EMA predictions, Mean-teacher model \cite{Tarvainen2017MeanTA} uses directly the EMA weights from the student to supervise the student under a consistency constraint. Authors of Dual student \cite{Ke2019DualSB} point out that the Mean Teacher converging to student along with training (coupling problem) prevents the teacher-student from exploiting more meaningful information from data. Inspired by Deep Co-training \cite{qiao2018deep}, they propose to train two independent students on stable samples which have same predictions and enough large feature difference. However, in unsupervised cross-domain Re-ID, labeled source domain and unlabeled target domain do not share the same identity classes, which makes traditional close-set SSL methods hard to use.

\paragraph{Fully unsupervised Re-ID.} Recently, several fully unsupervised Re-ID methods are proposed to further minimize the supervision, which does not require any Re-ID annotation. A bottom-up clustering framework is proposed in BUC \cite{Lin2019ABC}, which trains a network based on the clustering-based pseudo labels in an iterative way. \cite{Lin2020UnsupervisedPR} replaces clustering-based pseudo labels with similarity-based softened labels. Different to image-based unsupervised Re-ID, \cite{Wu2020TrackletSL} learns tacklet information with clustering-based pseudo labels. MMT \cite{ge2020mutual} can be transferred into an unsupervised method by removing the pre-training in source domain. However, without different source domain initializations, divergence between peer networks can not be enough encouraged in MMT. Instead of different source domain initializations, divergence is encouraged by asymmetric network structures, which is more suitable for fully unsupervised Re-ID.


\section{Proposed Method}

\subsection{Overview}
Given two datasets: one labeled source dataset $D_s$ and one unlabeled target dataset $D_t$, the objective of UDA is to adapt a source pretrained model $M_{pre}$ to the target dataset with unlabeled target data. To achieve this goal, we propose a two-staged adaptation approach based on Mean Teacher Model. We focus on the coupling problem (teacher and student converge to each other) existing inside the original Mean Teacher. Asymmetric branches and cross-branch supervision are proposed in this paper to address this problem and to enhance the diversity in the network, which show great effectiveness for UDA Re-ID. 

\begin{figure}
\begin{center}
   \includegraphics[width=1\linewidth]{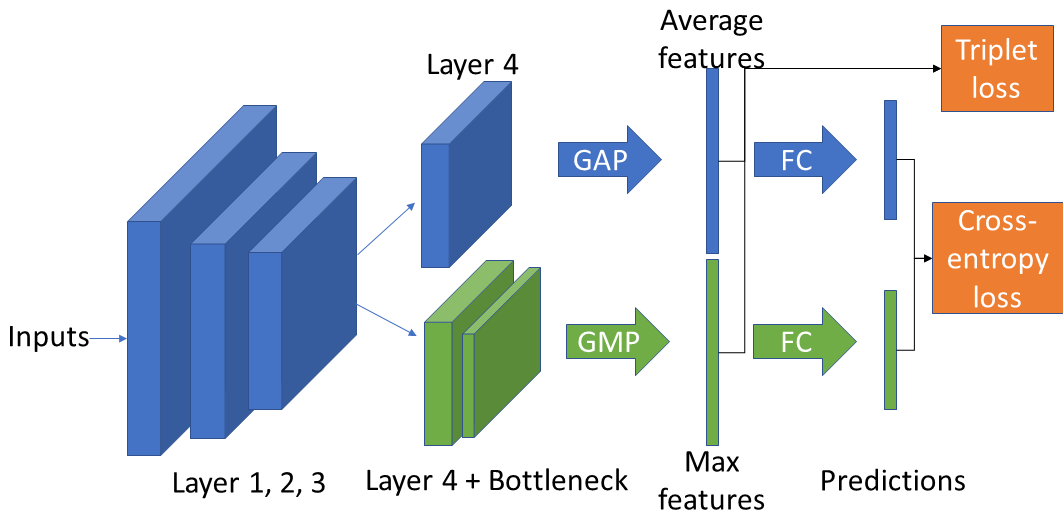}
\end{center}
   \caption{Source domain pre-training for asymmetric branched network. One ResNet bottleneck block corresponds to three convolutional layers. For UDA setting, inputs are labelled images from source training set. GAP refers Global Average Pooling, while GMP refers to Global Max Pooling. FC refers to Fully Connected layer.}
\label{fig:figure2}
\end{figure}

\subsection{Asymmetric branches}
A multi-branch structure is widely used in the fully supervised Re-ID methods, especially in global-local feature based methods \cite{Fu2018HorizontalPM,Dai2018BatchDN,Chen_2020_WACV}. Such structure keeps independence between branches, which makes features extracted from different branches diversified. In the unsupervised Re-ID, we conduct clustering on appearance signatures computed from person images to generate pseudo labels. 
The quality of appearance signatures can be improved by
extracting distinct meaningful features from different branches. Thus, we duplicate last layers of a backbone network and make them different in the structure, which we call Asymmetric Branches.

Asymmetric branches are illustrated in Figure \ref{fig:figure2}. For a ResNet-based \cite{He2015DeepRL} backbone, the layer 4 is duplicated. The first branch is kept unchanged as the one used in the original backbone: 3 bottlenecks and global average pooling (GAP). The second branch is composed of 4 bottlenecks and global max pooling (GMP). The GAP perceives global information, while the GMP focuses on the most discriminative information (most distinguishable identity information, such as a red bag or a yellow t-shirt). Asymmetric branches improve appearance signature quality by enhancing the feature diversity, which is validated by source pre-training performance boost in Table \ref{table:2} as well as examples in Figure \ref{fig:figure4}. They further improve the quality of pseudo labels during the adaptation, which is validated by target adaptation performance in Table \ref{table:2}.

\subsection{Asymmetric Branched Mean Teaching}
 We call our proposed adaptation method Asymmetric Branched Mean Teaching (ABMT). Our proposed ABMT contains two stages: Source pre-training and Target adaptation. 
 

  
      
      
    
    
      

\begin{figure*}
\begin{center}
   \includegraphics[width=0.8\linewidth]{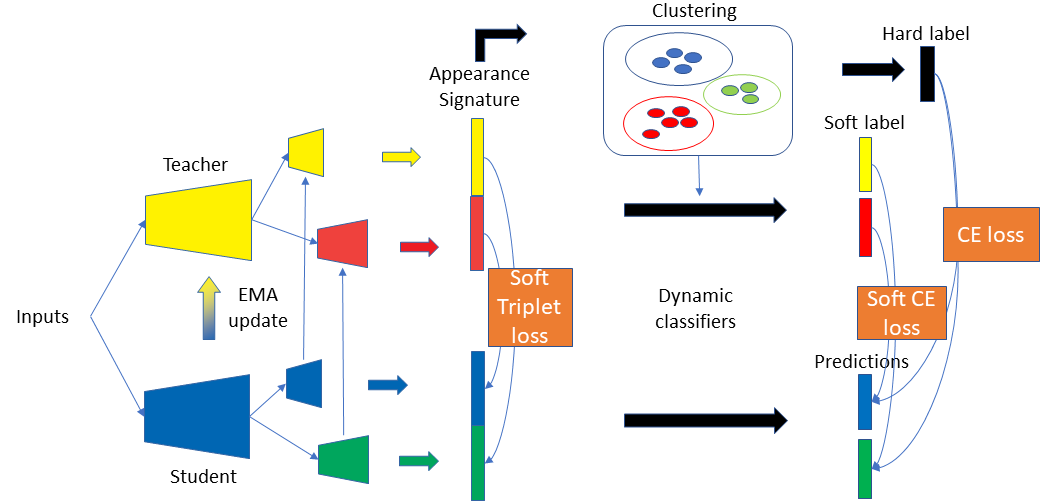}
\end{center}
   \caption{ABMT adaptation. For UDA setting, inputs are training set images from both source and target domains. For fully unsupervised setting, inputs are unlabeled images from target training set. }
\label{fig:figure3}
\end{figure*}

\subsubsection{Source domain supervised pre-training}
In the first stage, we train a network in the fully supervised way on the source domain. Thanks to this stage, the model used for adaptation obtains a basic Re-ID capacity, which helps to alleviate pseudo label noise. Given a source sample $x^s_i$ and its ground truth identity $y^\prime_i$, the network (with weight $\theta$) encodes $x^s_i$ into average $F_{a}(x^s_i|\theta)$ and max features $F_{m}(x^s_i|\theta)$ and then gets two predictions $P_{a}(x^s_i|\theta)$ and $P_{m}(x^s_i|\theta)$. Cross-entropy $L_{ce}$ and batch hard triplet \cite{hermans2017defense} $L_{tri}$ losses are used in this stage as shown in Figure \ref{fig:figure2}.


The whole network is trained with a combination of both losses:
\begin{equation}
\begin{aligned}
L_{scr} = & \lambda_{ce}^{s}L_{ce}(P_{a}(x^s_i|\theta),y^\prime_i) + \lambda_{ce}^{s}L_{ce}(P_{m}(x^s_i|\theta),y^\prime_i)+\\
      &\lambda_{tri}^{s}L_{tri}(F_{a}(x^s_i|\theta),y^\prime_i)
       +\lambda_{tri}^{s}L_{tri}(F_{m}(x^s_i|\theta),y^\prime_i)
\end{aligned}
\label{equ:3}
\end{equation}
\subsubsection{Target domain unsupervised adaptation}
The adaptation procedure is illustrated in Figure \ref{fig:figure3}. It contains two components: Clustering-based hard label generation and Cross-branch teacher-based soft label training. After adaptation, only teacher network is used during the inference.

\paragraph{Clustering-based hard label generation.}
In previous UDA Re-ID methods, distance-based K-Means \cite{ge2020mutual} and density-based clustering DBSCAN \cite{yang2020asy,song2020unsupervised} are main approaches to generate pseudo labels. 

We follow the state-of-the-art DBSCAN based clustering method presented in \cite{song2020unsupervised}. To adapt it to our proposed asymmetric branches, we concatenate the average and max features from asymmetric branches in the teacher network as appearance signatures. Images belonging to the same identity should have the same nearest neighbors in the feature space. Distance metric for DBSCAN are obtained by k-reciprocal re-ranking encoding \cite{zhong2017re} between target domain and source domain samples. 

The density-based clustering generates unfixed cluster numbers at different epochs, which means old classifiers from the last epoch can not be reused after a new clustering. Thus, we simply create new classifiers depending on the number of clusters at the beginning of each epoch. We take normalized mean features of each cluster from the average branch to initialize the average branch classifiers and similarly normalized max features from max branch to initialize the max branch classifiers. We call these classifiers with flexible dimension "Dynamic Classifiers". With the help of these Dynamic Classifiers, the student is trained on cluster components (outliers are discarded) with cross-entropy loss:
\begin{equation}
   L_{ce} = - \sum_{i} ({y_i' \log(P_{m}(x^t_i|\theta)) }) - \sum_{i} ({y_i' \log(P_{a}(x^t_i|\theta)) })
\label{equ:4}
\end{equation}
where $y_i'$ is the clustering based hard label and $P_{a}(x^t_i|\theta)$ and $P_{m}(x^t_i|\theta)$ are student predictions from both asymmetric branches.

\paragraph{Cross-branch teacher-based soft label training.} Clustering algorithms generate hard pseudo labels whose confidences are 100\%. Since Re-ID is a fine-grained recognition problem, people with similar clothes are not rare in the dataset. Hard pseudo labels of these similar samples can be extremely noisy. In this case, soft pseudo labels (confidences $<$ 100\%) are more reliable. Learning with both hard and soft pseudo labels can effectively alleviate label noise.

The Mean Teacher Model \cite{Tarvainen2017MeanTA} (teacher weights $\theta^{\prime}$) uses the EMA weights of the student model (student weights $\theta$). The Mean Teacher Model shows strong capacity to handle label noise and avoids error amplification along with training.
We define $\theta_t^{\prime}$ at training step t as the EMA of successive  weights:
\begin{equation}
\theta_t^{\prime} =
\begin{cases}
    \theta_t,& \text{if } t=0\\
     \alpha\theta_{t-1}^{\prime}+(1-\alpha)\theta_t, & \text{otherwise}
\end{cases}
\label{equ:5}
\end{equation}
where $\alpha$ is a smoothing coefficient that controls the self-ensembling speed of the Mean Teacher.

Despite these advantages of Mean Teacher, such self-ensembling teacher-student networks (the teacher is formed by EMA weights of the student, and the student is supervised by the teacher) face the coupling problem. We use the Mean Teacher soft label generator as in \cite{ge2020mutual} and address the coupling problem by cross-branch supervision. Each branch in the student is supervised by a teacher branch which has different structure. Weight diversity between the paired teacher-student can be better kept. Given one target domain sample $x^t_i$, the teacher (teacher weights $\theta^{\prime}$) encodes it into two feature vectors from two asymmetric branches, average features $F_{a}(x^t_i|\theta^{\prime})$ and max features $F_{m}(x^t_i|\theta^{\prime})$. The dynamic classifiers then transform these two feature vectors into two predictions respectively $P_{a}(x^t_i|\theta^{\prime})$ and $P_{m}(x^t_i|\theta^{\prime})$. Similarly, features of the student (student weights $\theta$) are $F_{a}(x^t_i|\theta)$ and $F_{m}(x^t_i|\theta)$, while predictions are $P_{a}(x^t_i|\theta)$ and $P_{m}(x^t_i|\theta)$. The predictions from the teacher supervise those from the student with a soft cross-entropy loss \cite{Hinton2015DistillingTK} in a cross-branch manner, which can be formulated as 
\begin{gather}
L_{sce}^{a \to m} = - \sum_{i} ({P_{a}(x^t_i|\theta^{\prime}) \log(P_{m}(x^t_i|\theta))}) \label{6}\\
L_{sce}^{m \to a} = - \sum_{i} ({P_{m}(x^t_i|\theta^{\prime}) \log(P_{a}(x^t_i|\theta))}) \label{7}
\end{gather}  
To further enhance the teacher-student networks' discriminative capacity, the features in the teacher supervise those of the student with a soft triplet loss \cite{ge2020mutual}:
\begin{gather}
L_{stri}^{a \to m} = - \sum_{i} ({T_{a}(x^t_i|\theta^{\prime}) \log(T_{m}(x^t_i|\theta))}) \label{8}\\
L_{stri}^{m \to a} =  - \sum_{i} ({T_{m}(x^t_i|\theta^{\prime}) \log(T_{a}(x^t_i|\theta))}) \label{9}
\end{gather}  
where \small $T(x^t_i|\theta)=\frac{exp(\left\lVert F(x^t_i|\theta) - F(x^t_p|\theta)\right\rVert_2 )}{exp(\left\lVert F(x^t_i|\theta) - F(x^t_p|\theta)\right\rVert_2 )+exp(\left\lVert F(x^t_i|\theta) - F(x^t_n|\theta)\right\rVert_2)}$ \normalsize
is the softmax triplet distance of the sample $x^t_i$, its hardest positive $x^t_p$ and its hardest negative $x^t_n$ in a mini-batch. By minimizing the soft triplet loss, the softmax triplet distance in a mini-batch from the student is encouraged to get as close as possible to the distance from the teacher. The positive and negative samples within a mini-batch are decided by clustering-based hard pseudo labels. It can effectively improve the UDA Re-ID performance. The teacher-student networks are trained end-to-end with Equation \eqref{equ:4}, \eqref{6}, \eqref{7}, \eqref{8}, \eqref{9}.
\begin{equation}
\begin{aligned}
L_{target} =  &\lambda_{ce}^{t} L_{ce} + \lambda_{sce}^{t}(L_{sce}^{a \to m} + L_{sce}^{m \to a})\\
 &+ \lambda_{stri}^{t}(L_{stri}^{a \to m} + L_{stri}^{m \to a})
\end{aligned}
\label{equ:10}
\end{equation}

\begin{figure*}
\begin{center}
   \includegraphics[width=0.8\linewidth]{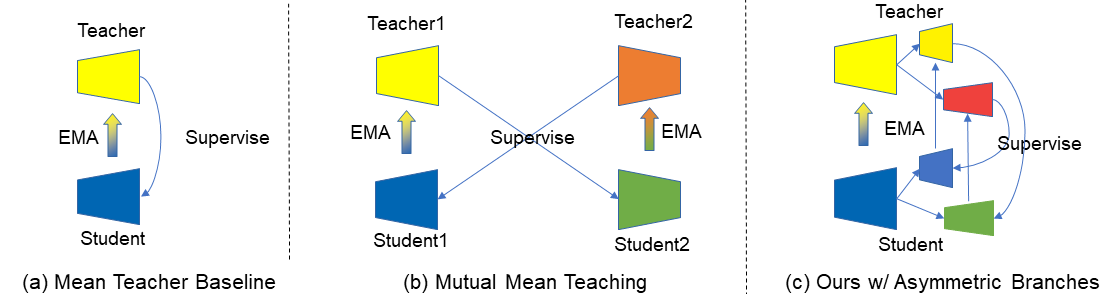}
\end{center}
   \caption{Comparison between (a) Mean Teacher Baseline (b) Mutual Mean Teaching \cite{ge2020mutual} and (c) our Mean Teacher with cross-branch supervised asymmetric branches. Teacher network is formed by exponential moving average (EMA) values of student network. }
\label{fig:figure1}
\end{figure*}

\section{Coupling Problem in Mean Teacher Based Methods  }
The Mean Teacher Baseline is illustrated in Figure \ref{fig:figure1} (a) where the student gets supervision from its own EMA weights. In the Mean Teacher Baseline, the student and the teacher quickly converge to each other (coupling problem), which prevents them from exploring more diversified information. Authors of MMT \cite{ge2020mutual} propose to pre-train 2 student networks with different seeds. As illustrated in Figure \ref{fig:figure1} (b), two Mean Teacher networks are formed separately from two students, which alleviates the coupling problem. However, different initializations decouple both teacher peers only at first epochs. Without a diversity encouragement during the adaptation, both teachers still converge to each other along with training. In Figure \ref{fig:figure1} (c), our proposed asymmetric branches provide a diversity encouragement during the adaptation, which decouples both teacher peers at all epochs. 

\begin{figure*}
\begin{center}
   \includegraphics[width=1\linewidth]{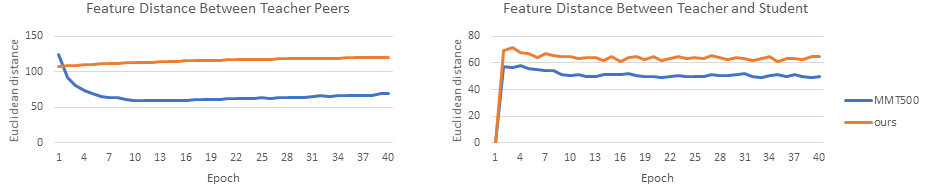}
\end{center}
   \caption{Distance comparison between features extracted from a ResNet50 backbone on all samples in DukeMTMC-reid training set for Market $\to$ Duke task. \textbf{Left}: Feature distance between two teacher models in MMT and between two teacher branches in our proposed method. \textbf{Right}: Feature distance between teacher and student networks.}
\label{fig:figure5}
\end{figure*}

To validate our idea, we propose to measure Euclidean distance of appearance signature features between two teacher networks or two teacher branches. We extract feature vectors after global pooling on all images in the target training set. Then, we calculate the Euclidean distance between feature vectors of both teachers and sum up the distance of every image as the final feature distance. If the feature distance is large, we can say that both teacher peers extract diversified features. Otherwise, the teacher peers converge to each other. As we can see from the left curves in Figure \ref{fig:figure5}, the feature distance between two teachers in MMT is large at the beginning, but it decreases and then stabilizes. Differently, the feature distance between two branches in our proposed method remains large during the training. Moreover, we visualize the Euclidean distance of appearance signature features on all target training samples between teacher and student networks in Figure \ref{fig:figure5} right curves. Our method can maintain a larger distance, which shows that it can better decouple teacher-student networks.

\section{Experiments}
\subsection{Datasets and Evaluation Protocols}
Our proposed adaptation method is evaluated on 3 Re-ID datasets: Market-1501, DukeMTMC-reID and MSMT17. \textbf{Market-1501} \cite{Zheng2015ScalablePR} dataset is collected in front of a supermarket in Tsinghua University from 6 cameras. It contains 12,936 images of 751 identities in the training set and 19,732 images of 750 identities in the testing set. \textbf{DukeMTMC-reID} \cite{ristani2016MTMC} is a subset of the DukeMTMC dataset. It contains 16,522 images of 702 persons in the training set, 2,228 query images and 17,661 gallery images of 702 persons for testing from 8 cameras. \textbf{MSMT17} \cite{wei2018person} is a large-scale Re-ID dataset, which contains 32,621 training images of 1,041 identities and 93,820 testing images of 3,060 identities collected from 15 cameras. Both Cumulative Matching Characteristics (CMC) and mean Average Precisions (mAP) are used in our experiments. 

\begin{table*}
\begin{center}
\scalebox{0.85}{
\begin{tabular}{l|cc|cc|cc|cc}
\hline
\multirow{2}{*}{UDA Methods}  & \multicolumn{2}{|c|}{Market $\to$ Duke } & \multicolumn{2}{|c|}{Duke $\to$ Market}& \multicolumn{2}{|c|}{Market $\to$ MSMT } & \multicolumn{2}{|c}{Duke $\to$ MSMT} \\ \cline{2-9}
\multicolumn{1}{c|}{} & \multicolumn{1}{c}{mAP} & \multicolumn{1}{c|}{Rank1} & \multicolumn{1}{c}{mAP} & \multicolumn{1}{c|}{Rank1} & \multicolumn{1}{c}{mAP} & \multicolumn{1}{c|}{Rank1} & \multicolumn{1}{c}{mAP} & \multicolumn{1}{c}{Rank1} \\ 
\hline\hline
HHL (ECCV'18)\cite{Zhong_2018_ECCV}&27.2&46.9&31.4&62.2&-&-&-&-\\
ECN (CVPR'19)\cite{zhong2019invariance}&40.4&63.3&43.0&75.1&8.5&25.3&10.2&30.2\\
PCB-PAST (ICCV'19)\cite{Zhang2019SelfTrainingWP}&54.3&72.4&54.6&78.4&-&-&-&-\\
SSG (ICCV'19)\cite{Fu2018SelfSimilarityGA}&53.4&73.0&58.3&80.0&13.2&31.6&13.3&32.2\\
UDAP (PR'20)\cite{song2020unsupervised}&49.0&68.4&53.7&75.8&-&-&-&-\\
ACT (AAAI'20)\cite{yang2020asy}&54.5&72.4&60.6&80.5&-&-&-&-\\
ECN+ (PAMI'20) \cite{zhong2020learning}&54.4&74.0&63.8&84.1&15.2&40.4&16.0&42.5\\
MMT500 (ICLR'20)(ResNet50)\cite{ge2020mutual}&63.1&76.8&71.2&87.7&16.6&37.5&17.9&41.3\\
MMT700 (ICLR'20)(ResNet50)\cite{ge2020mutual}&65.1&78.0&69.0&86.8&-&-&-&-\\
MMT1500 (ICLR'20)(ResNet50)\cite{ge2020mutual}&-&-&-&-&22.9&49.2&23.3&50.1\\
\hline
ours (ResNet50)&\textbf{69.1}&\textbf{82.0}&\textbf{78.3}&\textbf{92.5}&\textbf{23.2}&\textbf{49.2}&\textbf{26.5}&\textbf{54.3}\\
\hline\hline
MMT500 (ICLR'20)(IBN-ResNet50)\cite{ge2020mutual}&65.7&79.3&76.5&90.9&19.6&43.3&23.3&50.0\\
MMT700 (ICLR'20)(IBN-ResNet50)\cite{ge2020mutual}&68.7&81.8&74.5&91.1&-&-&-&-\\
MMT1500 (ICLR'20)(IBN-ResNet50)\cite{ge2020mutual}&-&-&-&-&26.6&54.4&29.3&58.2\\
\hline
ours (IBN-ResNet50)&\textbf{70.8}&\textbf{83.3}&\textbf{80.4}&\textbf{93.0}&\textbf{27.8}&\textbf{55.5}&\textbf{33.0}&\textbf{61.8}\\
\hline
\end{tabular}}
\end{center}
\caption{Comparison of unsupervised domain adaptation (UDA) Re-ID methods (\%) on medium-to-medium datasets (Market$\to$ Duke and Duke $\to$ Market) and medium-to-large datasets (Market $\to$ MSMT and Duke $\to$ MSMT). }
\label{table:1}
\end{table*}

\subsection{Implementation details}
Hyper-parameters used in our proposed method are searched empirically from the Market$\to$ Duke task and kept the same for the other tasks. To conduct fair comparison with state-of-the-arts, we use a ImageNet \cite{deng2009imagenet} pre-trained ResNet-50 \cite{He2015DeepRL} as our backbone network. The backbone can be extended to ResNet-based networks designed for cross domain tasks, \eg, IBN-ResNet-50 \cite{pan2018two}. An Adam optimizer with a weight decay rate of 0.0005 is used to optimize our networks. Our networks are trained on 4 Nvidia 1080Ti GPUs under Pytorch \cite{Paszke2019PyTorchAI} framework. Detailed configurations are given in the following paragraphs.

\paragraph{Stage1: Source domain supervised pre-training.} We set $\lambda_{ce}^{s}=0.5$ and $\lambda_{tri}^{s}=0.5$ in Equation \ref{equ:3}. The max epoch $E_{pre}$ is set to 80. For each epoch, the networks are trained $R_{pre}=200$ iterations.The initial learning rate is set to 0.00035 and is multiplied by 0.1 at the 40th and 70th epoch. For each iteration, 64 images of 16 identities are resized to 256*128 and fed into networks.

\paragraph{Stage2: Target domain unsupervised adaptation.}
For the clustering, we set the minimum cluster samples to 4 and the density radius $r$=0.002. Re-ranking parameters for calculating distances are kept the same as in \cite{song2020unsupervised} for UDA setting. Re-ranking between source and target domain is not considered for fully unsupervised setting. The Mean Teacher network is initialized and updated in the way of Equation \ref{equ:5} with a smoothing  coefficient $\alpha=0.999$. We set $\lambda_{ce}^{t}=0.5$, $\lambda_{sce}^{t}=0.5$ and $\lambda_{stri}^{t}=1$ in Equation \ref{equ:10}. The adaptation epoch $E_{ada}$ is set to 40. For each epoch, the networks are trained $R_{ada}=400$ iterations with a fixed learning rate 0.00035. For each iteration, 64 images of 16 clustering-based pseudo identities are resized to 256*128 and fed into networks with Random erasing \cite{Zhong2020RandomED} data augmentation.  

\subsection{Comparison with State-of-the-Art Methods}
\begin{figure}
\begin{center}
   \includegraphics[width=1\linewidth]{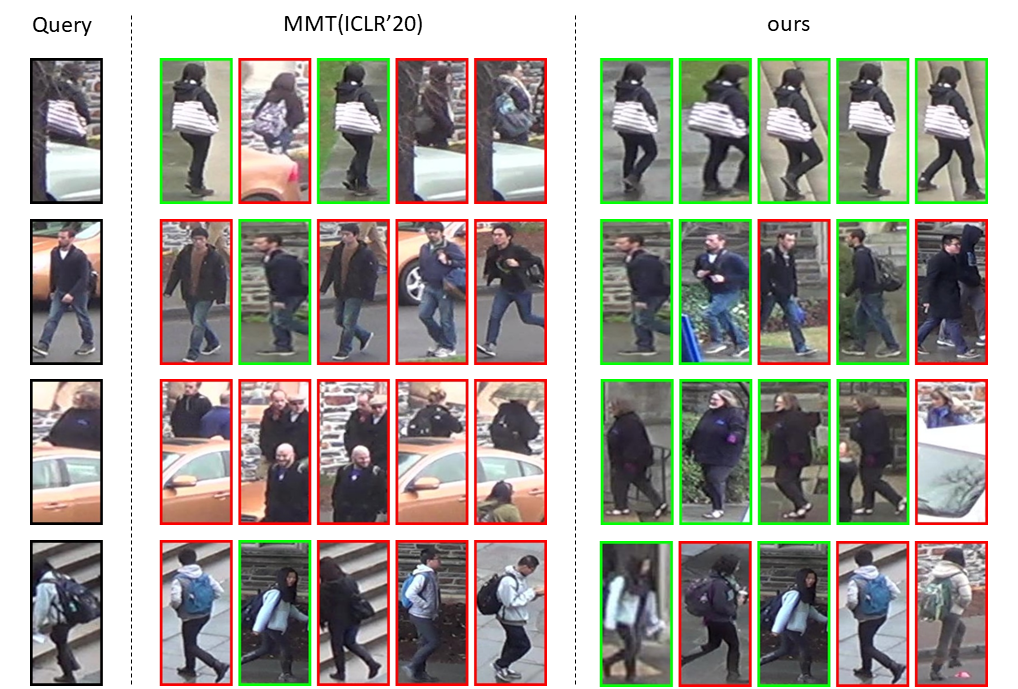}
\end{center}
   \caption{Examples of retrieved most similar 5 images in Market $\to$ Duke task from MMT \cite{ge2020mutual} and our proposed method. Given a query image, different identity images are highlighted by red bounding boxes, while same identity images are highlighted by green bounding boxes.}
\label{fig:figure4}
\end{figure}

We compare our proposed methods with state-of-the-art UDA methods in Table \ref{table:1} for 4 cross-dataset Re-ID tasks: Market$\to$ Duke, Duke $\to$ Market, Market $\to$ MSMT and Duke $\to$ MSMT. Post-processing techniques (\eg, Re-ranking \cite{zhong2017re}) are not used in the comparison. 
Our proposed method outperforms MMT \cite{ge2020mutual} (cluster number is set to 500, 700 and 1500 respectively). We can also adjust the density radius in DBSCAN depending on target domain size to get a better performance, but we think it is hard to know the target domain size in the real world. With an IBN-ResNet50 \cite{pan2018two} backbone, the performance on 4 tasks can be further improved. Examples of retrieved images are illustrated in Figure \ref{fig:figure4}. Compared to MMT, embeddings from our proposed method contain more discriminative appearance information (\eg, shoulder bag in the first row), which are robust to noisy information (\eg, pose variation in the second row, occlusion in the third row and background variation in the fourth row). This qualitative comparison confirms that appearance signatures of our proposed method are of improved quality.

\begin{table}
\begin{center}
\scalebox{0.8}{
\begin{tabular}{l|cc|cc}
\hline
\multirow{2}{*}{Unsupervised methods}  & \multicolumn{2}{|c}{Market} & \multicolumn{2}{|c}{Duke} \\ \cline{2-5}
\multicolumn{1}{c|}{} & \multicolumn{1}{c}{mAP} & \multicolumn{1}{c|}{Rank1} & \multicolumn{1}{c}{mAP} & \multicolumn{1}{c}{Rank1} \\ 
\hline
MMT500*(ICLR'20)\cite{ge2020mutual} &26.9&48.0&7.3&12.7\\
BUC (AAAI'19)\cite{Lin2019ABC}  &30.6&61.0&21.9&40.2\\ 
SoftSim (CVPR'20)\cite{Lin2020UnsupervisedPR} &37.8&71.7&28.6&52.5\\
TSSL (AAAI'20)\cite{Wu2020TrackletSL} &43.3&71.2&38.5&62.2\\
MMT*+DBSCAN (ICLR'20)\cite{ge2020mutual}&53.5&73.1&54.5&69.5\\
\hline
ours w/o Source pre-training&\textbf{65.1}&\textbf{82.6}&\textbf{63.1}&\textbf{77.7}\\
\hline
\end{tabular}}
\end{center}
\caption{Comparison of unsupervised Re-ID methods (\%) with a ResNet50 backbone on Market and Duke datasets. * refers to our implementation where we remove the source pre-training step. DBSCAN refers to a DBSCAN clustering based on re-ranked distance.}
\label{table:unsupervised}
\end{table}

\begin{table*}
\begin{center}
\scalebox{0.85}{
\begin{tabular}{l|cc|cc}
\hline
\multirow{2}{*}{Source pre-training}  & \multicolumn{2}{|c|}{Market $\to$ Duke } & \multicolumn{2}{|c}{Duke $\to$ Market} \\ \cline{2-5}
\multicolumn{1}{c|}{} & \multicolumn{1}{c}{mAP} & \multicolumn{1}{c|}{Rank1} & \multicolumn{1}{c}{mAP} & \multicolumn{1}{c}{Rank1} \\ 
\hline
ResNet50&29.6&46.0&31.8&61.9\\
ResNet50+AB  &31.5&49.7&33.2&63.2\\
\hline\hline
\multirow{2}{*}{Target adaptation}  & \multicolumn{2}{|c|}{Market $\to$ Duke } & \multicolumn{2}{c}{Duke $\to$ Market} \\ \cline{2-5}
\multicolumn{1}{c|}{} & \multicolumn{1}{c}{mAP} & \multicolumn{1}{c|}{Rank1} & \multicolumn{1}{c}{mAP} & \multicolumn{1}{c}{Rank1} \\ 
\hline
MT-Baseline+K-Means &59.9&74.8&68.9&88.2\\
MT-Baseline+DBSCAN &61.9&77.3&69.9&88.3\\
MT-Baseline+K-Means+AB &64.7&78.1&74.8&90.5\\
MT-Baseline+K-Means+AB+Cross-branch &66.4&79.9&76.8&91.7\\
MT-Baseline+DBSCAN+AB &67.8&81.1&77.3&92.0\\
ABMT(MT-Baseline+DBSCAN+AB+Cross-branch) &\textbf{69.1}&\textbf{82.0}&\textbf{78.3}&\textbf{92.5}\\
\hline
ABMT+Stochastic data augmentation&68.8&81.2&77.6&91.7\\
ABMT+Drop out&68.3&81.8&77.9&92.0\\
\hline
\end{tabular}}
\end{center}
\caption{Ablation studies with ResNet50 backbone. MT-Baseline corresponds to the Mean Teacher Baseline in Figure \ref{fig:figure1} (a) with a ResNet-50. K-Means refers to a K-Means++ clustering whose cluster number is set to 500. AB refers to asymmetric branches. DBSCAN refers to a DBSCAN clustering \cite{Ester1996ADA}. }
\label{table:2}
\end{table*}

\begin{table}
\begin{center}
\scalebox{0.85}{
\begin{tabular}{l|cc|cc}
\hline
\multirow{2}{*}{Structure}  & \multicolumn{2}{|c|}{Market $\to$ Duke } & \multicolumn{2}{c}{Duke $\to$ Market} \\ \cline{2-5}
\multicolumn{1}{c|}{} & \multicolumn{1}{c}{mAP} & \multicolumn{1}{c|}{Rank1} & \multicolumn{1}{c}{mAP} & \multicolumn{1}{c}{Rank1} \\ 
\hline
ABMT &\textbf{69.1}&\textbf{82.0}&\textbf{78.3}&\textbf{92.5}\\
ABMT w/o different pooling&65.2&79.7&74.2&90.1\\
ABMT w/o extra bottleneck&67.5&80.6&77.6&92.4\\
\hline
ABMT + one more branch&68.1&80.7&76.2&90.4\\
\hline
\end{tabular}}
\end{center}
\caption{Ablation studies on structure of asymmetric branches. }
\label{table:3}
\end{table}

\begin{table}
\begin{center}
\scalebox{0.85}{
\begin{tabular}{l|cc|cc}
\hline
\multirow{2}{*}{Loss}  & \multicolumn{2}{|c|}{Market $\to$ Duke } & \multicolumn{2}{c}{Duke $\to$ Market} \\ \cline{2-5}
\multicolumn{1}{c|}{} & \multicolumn{1}{c}{mAP} & \multicolumn{1}{c|}{Rank1} & \multicolumn{1}{c}{mAP} & \multicolumn{1}{c}{Rank1} \\ 
\hline
ABMT&\textbf{69.1}&\textbf{82.0}&\textbf{78.3}&\textbf{92.5}\\
ABMT w/o $L_{ce}$&52.5&69.6&57.5&79.8\\
ABMT w/o $L_{sce}$&66.7&79.8&77.7&92.2\\
ABMT w/o $L_{stri}$&64.7&78.5&75.5&91.2\\
\hline
\end{tabular}}
\end{center}
\caption{Ablation studies on loss functions. }
\label{table:4}
\end{table}

We compare unsupervised Re-ID methods in Table \ref{table:unsupervised}. Since the Mean Teacher is designed for handling label noise, it is interesting to see the performance without source pre-training, which introduces more label noise during the adaptation. This setting corresponds to an unsupervised Re-ID. We use ImageNet pretained weights as initialization. Our proposed method outperforms previous unsupervised Re-ID by a large margin, which shows that ImageNet initialization can provide basic discriminative capacity for Re-ID.

MMT \cite{ge2020mutual} is the first Mean Teacher based UDA Re-ID method. Authors of MMT propose to use 2 students and 2 teachers with different initialization and stochastic data augmentation to address the coupling problem. We also use Mean Teacher soft pseudo labels but propose a different decoupling solution. Features in asymmetric branches are always extracted in different manners during the adaptation. Compared to MMT, our proposed method has less parameters (approximately 10\% less parameters and 20\% less operations) but achieves better performance. Moreover, in the unsupervised scenario, we can not pre-train MMT with different seeds to obtain different Re-ID initializations. This decoupling strategy becomes inappropriate. Our decoupling strategy relies on structural asymmetry instead of different initializations, which is much more effective in the unsupervised scenario.

ACT \cite{yang2020asy} uses 2 networks, in which each network learns from its peer. Input data are split into inliers and ouliers after DBSCAN. Then, the first network selects small entropy inliers to train the second network, while the second selects small entropy outliers to train the first. This method enhances input asymmetry by data split. Differently, our proposed method focuses on neural network structure asymmetry. 

\subsection{Ablation Studies}

\paragraph{Effectiveness of each component in ABMT.} Compared with traditional clustering-based Re-ID methods, the performance improvement mainly comes from DBSCAN on re-ranked distance, asymmetric branches and cross-branch supervision. We use a Mean Teacher Baseline where original ResNet-50 and a K-Means++ clustering of 500 clusters are adopted. We conduct ablation studies by gradually adding one component at each time. Results are shown in Table \ref{table:2}. We can observe: (1) Our proposed asymmetric branches bring the most significant performance improvement during the adaptation. Moreover, as we can see from first two rows in Table \ref{table:2}, they can directly improve the domain generalizability of appearance signatures without target adaptation. (2) DBSCAN on re-ranked distance works better than a K-Means++ clustering of 500 clusters during the adaptation. (3) Cross-branch supervision works on asymmetric branches, which can further improve the adaptation performance.

\paragraph{Effectiveness of asymmetric branch structure.} To validate the effectiveness of our proposed asymmetric branch structure, we compare several possible structures: (1) 2 branches with different pooling methods and different depths, (2) 2 branches with same pooling methods (global average pooling) but different depths, (3) 2 branches with different methods but same depths, (4) 3 branches where the new branch is composed of 5 bottleneck blocks and global average pooling. From results given in Table \ref{table:3}, we can conclude that different pooling methods play a more important role in asymmetric branches. 

\paragraph{Effectiveness of loss functions.} We conduct ablation studies on loss functions used in our proposed method and report results in Table \ref{table:4}. The degree of influence of 3 loss functions used in our proposed method: $L_{ce} > L_{sce} > L_{stri}$.
\paragraph{Can traditional decoupling methods further improve the performance?} 
Stochastic data augmentation (teacher inputs and student inputs are under stochastic data augmentation methods) and drop out (teacher feature vectors and student feature vectors are under independent drop out operations before classifiers) are 2 widely-used methods to provide random noise, which also helps to decouple the weights between the teacher and the student. We conduct experiments with stochastic data augmentation
. The results in Table \ref{table:2} show that they can not further improve the UDA Re-ID performance. These methods are not designed for fine-grained Re-ID task. As UDA Re-ID performance is already very high, they can not contribute anymore.  

\section{Conclusion}
In this paper, we propose a novel unsupervised cross-domain Re-ID framework. Our proposed method is mainly based on learning from noisy pseudo labels generated by clustering and Mean Teacher. A self-ensembled Mean Teacher is robust to label noise, but the coupling problem inside paired teacher-student networks leads to a performance bottleneck. To address this problem, we propose asymmetric branches and cross-branch supervision, which can effectively enhance the diversity in two aspects: appearance signature features and teacher-student weights. By enhancing the diversity in the teacher-student networks, our proposed method achieves better performance on both unsupervised domain adaptation and fully unsupervised Re-ID tasks. 

{\small
\bibliographystyle{ieee_fullname}
\bibliography{egbib}

\begin{thebibliography}{10}\itemsep=-1pt

\bibitem{Chen_2020_WACV}
Hao Chen, Benoit Lagadec, and Francois Bremond.
\newblock Learning discriminative and generalizable representations by
  spatial-channel partition for person re-identification.
\newblock In {\em The IEEE Winter Conference on Applications of Computer Vision
  (WACV)}, March 2020.

\bibitem{chen2019instance}
Yanbei Chen, Xiatian Zhu, and Shaogang Gong.
\newblock Instance-guided context rendering for cross-domain person
  re-identification.
\newblock In {\em Proceedings of the IEEE International Conference on Computer
  Vision}, pages 232--242, 2019.

\bibitem{Dai2018BatchDN}
Zuozhuo Dai, Mingqiang Chen, Siyu Zhu, and Ping Tan.
\newblock Batch dropblock network for person re-identification and beyond.
\newblock {\em 2019 IEEE/CVF International Conference on Computer Vision
  (ICCV)}, pages 3690--3700, 2018.

\bibitem{deng2009imagenet}
Jia Deng, Wei Dong, Richard Socher, Li-Jia Li, Kai Li, and Li Fei-Fei.
\newblock Imagenet: A large-scale hierarchical image database.
\newblock In {\em 2009 IEEE conference on computer vision and pattern
  recognition}, pages 248--255. Ieee, 2009.

\bibitem{Ester1996ADA}
Martin Ester, Hans-Peter Kriegel, J{\"o}rg Sander, and Xiaowei Xu.
\newblock A density-based algorithm for discovering clusters in large spatial
  databases with noise.
\newblock In {\em KDD}, 1996.

\bibitem{Fu2018SelfSimilarityGA}
Yang Fu, Yunchao Wei, Guanshuo Wang, Xi Zhou, Honghui Shi, and Thomas~S. Huang.
\newblock Self-similarity grouping: A simple unsupervised cross domain
  adaptation approach for person re-identification.
\newblock {\em 2019 IEEE/CVF International Conference on Computer Vision
  (ICCV)}, pages 6111--6120, 2018.

\bibitem{Fu2018HorizontalPM}
Yang Fu, Yunchao Wei, Yuqian Zhou, Honghui Shi, Gao Huang, Xinchao Wang,
  Zhiqiang Yao, and Thomas~S. Huang.
\newblock Horizontal pyramid matching for person re-identification.
\newblock In {\em AAAI}, 2018.

\bibitem{ge2020mutual}
Yixiao Ge, Dapeng Chen, and Hongsheng Li.
\newblock Mutual mean-teaching: Pseudo label refinery for unsupervised domain
  adaptation on person re-identification.
\newblock In {\em International Conference on Learning Representations}, 2020.

\bibitem{Han2018CoteachingRT}
Bo Han, Quanming Yao, Xingrui Yu, Gang Niu, Miao Xu, Weihua Hu, Ivor Wai-Hung
  Tsang, and Masashi Sugiyama.
\newblock Co-teaching: Robust training of deep neural networks with extremely
  noisy labels.
\newblock In {\em NeurIPS}, 2018.

\bibitem{He2015DeepRL}
Kaiming He, Xiangyu Zhang, Shaoqing Ren, and Jian Sun.
\newblock Deep residual learning for image recognition.
\newblock {\em 2016 IEEE Conference on Computer Vision and Pattern Recognition
  (CVPR)}, pages 770--778, 2015.

\bibitem{hermans2017defense}
Alexander Hermans, Lucas Beyer, and Bastian Leibe.
\newblock In defense of the triplet loss for person re-identification.
\newblock {\em arXiv preprint arXiv:1703.07737}, 2017.

\bibitem{Hinton2015DistillingTK}
Geoffrey~E. Hinton, Oriol Vinyals, and Jeffrey Dean.
\newblock Distilling the knowledge in a neural network.
\newblock {\em ArXiv}, abs/1503.02531, 2015.

\bibitem{Ke2019DualSB}
Zhanghan Ke, Daoye Wang, Qiong Yan, Jimmy Ren, and Rynson W.~H. Lau.
\newblock Dual student: Breaking the limits of the teacher in semi-supervised
  learning.
\newblock {\em 2019 IEEE/CVF International Conference on Computer Vision
  (ICCV)}, pages 6727--6735, 2019.

\bibitem{Laine2016TemporalEF}
Samuli Laine and Timo Aila.
\newblock Temporal ensembling for semi-supervised learning.
\newblock {\em ArXiv}, abs/1610.02242, 2016.

\bibitem{Lin2018MultitaskMF}
Shan Lin, Haoliang Li, Chang-Tsun Li, and Alex~Chichung Kot.
\newblock Multi-task mid-level feature alignment network for unsupervised
  cross-dataset person re-identification.
\newblock In {\em BMVC}, 2018.

\bibitem{Lin2019ABC}
Yutian Lin, Xuanyi Dong, Liang Zheng, Yan Yan, and Yi Yang.
\newblock A bottom-up clustering approach to unsupervised person
  re-identification.
\newblock In {\em AAAI}, 2019.

\bibitem{Lin2020UnsupervisedPR}
Yutian Lin, Lingxi Xie, Yu Wu, Chenggang Yan, and Qi Tian.
\newblock Unsupervised person re-identification via softened similarity
  learning.
\newblock {\em ArXiv}, abs/2004.03547, 2020.

\bibitem{pan2018two}
Xingang Pan, Ping Luo, Jianping Shi, and Xiaoou Tang.
\newblock Two at once: Enhancing learning and generalization capacities via
  ibn-net.
\newblock In {\em Proceedings of the European Conference on Computer Vision
  (ECCV)}, pages 464--479, 2018.

\bibitem{Paszke2019PyTorchAI}
Adam Paszke, Sam Gross, Francisco Massa, Adam Lerer, James Bradbury, Gregory
  Chanan, Trevor Killeen, Zeming Lin, Natalia Gimelshein, Luca Antiga, Alban
  Desmaison, Andreas K{\"o}pf, Edward Yang, Zach DeVito, Martin Raison, Alykhan
  Tejani, Sasank Chilamkurthy, Benoit Steiner, Lu Fang, Junjie Bai, and Soumith
  Chintala.
\newblock Pytorch: An imperative style, high-performance deep learning library.
\newblock In {\em NeurIPS}, 2019.

\bibitem{qiao2018deep}
Siyuan Qiao, Wei Shen, Zhishuai Zhang, Bo Wang, and Alan Yuille.
\newblock Deep co-training for semi-supervised image recognition.
\newblock In {\em Proceedings of the european conference on computer vision
  (eccv)}, pages 135--152, 2018.

\bibitem{ristani2016MTMC}
Ergys Ristani, Francesco Solera, Roger Zou, Rita Cucchiara, and Carlo Tomasi.
\newblock Performance measures and a data set for multi-target, multi-camera
  tracking.
\newblock In {\em European Conference on Computer Vision workshop on
  Benchmarking Multi-Target Tracking}, 2016.

\bibitem{song2020unsupervised}
Liangchen Song, Cheng Wang, Lefei Zhang, Bo Du, Qian Zhang, Chang Huang, and
  Xinggang Wang.
\newblock Unsupervised domain adaptive re-identification: Theory and practice.
\newblock {\em Pattern Recognition}, 102:107173, 2020.

\bibitem{Tarvainen2017MeanTA}
Antti Tarvainen and Harri Valpola.
\newblock Mean teachers are better role models: Weight-averaged consistency
  targets improve semi-supervised deep learning results.
\newblock In {\em NIPS}, 2017.

\bibitem{Wang2018TransferableJA}
Jingya Wang, Xiatian Zhu, Shaogang Gong, and Wei Li.
\newblock Transferable joint attribute-identity deep learning for unsupervised
  person re-identification.
\newblock {\em 2018 IEEE/CVF Conference on Computer Vision and Pattern
  Recognition}, pages 2275--2284, 2018.

\bibitem{wei2018person}
Longhui Wei, Shiliang Zhang, Wen Gao, and Qi Tian.
\newblock Person transfer gan to bridge domain gap for person
  re-identification.
\newblock In {\em Proceedings of the IEEE conference on computer vision and
  pattern recognition}, pages 79--88, 2018.

\bibitem{Wu2020TrackletSL}
Guile Wu, Xiatian Zhu, and Shaogang Gong.
\newblock Tracklet self-supervised learning for unsupervised person
  re-identification.
\newblock In {\em AAAI 2020}, 2020.

\bibitem{yang2020asy}
Fengxiang Yang, Ke Li, Zhun Zhong, Zhiming Luo, Xing Sun, Hao Cheng, Xiaowei
  Guo, Feiyue Huang, Rongrong Ji, and Shaozi Li.
\newblock Asymmetric co-teaching for unsupervised cross domain person
  re-identification.
\newblock 2020.

\bibitem{Yu2019HowDD}
Xingrui Yu, Bo Han, Jiangchao Yao, Gang Niu, Ivor Wai-Hung Tsang, and Masashi
  Sugiyama.
\newblock How does disagreement help generalization against label corruption?
\newblock In {\em ICML}, 2019.

\bibitem{Zhang2019SelfTrainingWP}
Xinyu Zhang, Jiewei Cao, Chunhua Shen, and Mingyu You.
\newblock Self-training with progressive augmentation for unsupervised
  cross-domain person re-identification.
\newblock {\em 2019 IEEE/CVF International Conference on Computer Vision
  (ICCV)}, pages 8221--8230, 2019.

\bibitem{Zheng2015ScalablePR}
Liang Zheng, Liyue Shen, Lu Tian, Shengjin Wang, Jingdong Wang, and Qi Tian.
\newblock Scalable person re-identification: A benchmark.
\newblock {\em 2015 IEEE International Conference on Computer Vision (ICCV)},
  pages 1116--1124, 2015.

\bibitem{zhong2017re}
Zhun Zhong, Liang Zheng, Donglin Cao, and Shaozi Li.
\newblock Re-ranking person re-identification with k-reciprocal encoding.
\newblock In {\em Proceedings of the IEEE Conference on Computer Vision and
  Pattern Recognition}, pages 1318--1327, 2017.

\bibitem{Zhong2020RandomED}
Zhun Zhong, Liang Zheng, Guoliang Kang, Shaozi Li, and Yi Yang.
\newblock Random erasing data augmentation.
\newblock In {\em AAAI}, 2020.

\bibitem{Zhong_2018_ECCV}
Zhun Zhong, Liang Zheng, Shaozi Li, and Yi Yang.
\newblock Generalizing a person retrieval model hetero- and homogeneously.
\newblock In {\em The European Conference on Computer Vision (ECCV)}, September
  2018.

\bibitem{zhong2019invariance}
Zhun Zhong, Liang Zheng, Zhiming Luo, Shaozi Li, and Yi Yang.
\newblock Invariance matters: Exemplar memory for domain adaptive person
  re-identiﬁcation.
\newblock In {\em Proceedings of IEEE Conference on Computer Vision and Pattern
  Recognition (CVPR)}, 2019.

\bibitem{zhong2020learning}
Zhun Zhong, Liang Zheng, Zhiming Luo, Shaozi Li, and Yi Yang.
\newblock Learning to adapt invariance in memory for person re-identification.
\newblock {\em IEEE Transactions on Pattern Analysis and Machine Intelligence},
  2020.

\end{thebibliography}
}

\end{document}